%% file: iclr2027_conference.tex
\documentclass{article} 
\usepackage{iclr2027_conference,times}

\input{math_commands.tex}

\usepackage{hyperref}
\usepackage{url}
\usepackage{times}
\usepackage{latexsym}
\usepackage{threeparttable}
\usepackage{colortbl}
\definecolor{oursbg}{RGB}{230, 245, 230}
\definecolor{retcolor}{RGB}{34, 139, 34}
\usepackage[dvipsnames]{xcolor}
\usepackage{adjustbox}
\usepackage{hyperref}
\usepackage{url}
\usepackage{booktabs}
\usepackage{amsfonts}
\usepackage{xcolor}
\usepackage{multirow}
\usepackage{tabularx}
\usepackage{graphicx}
\usepackage{amsmath}
\usepackage{amssymb}
\usepackage{algorithm}
\usepackage{algorithmic}
\usepackage{float}
\usepackage{fontawesome5}

\title{Coarse Indexing, Fine Evidence: Decoupling Temporal Granularity in Long-Video RAG}

\author{
\parbox{\textwidth}{\centering
  Zhe Jin$^{1,\ast}$,
  Zhimin Lin$^{2,\ast}$,
  Bin Zheng$^{1}$,
  Junhua Fang$^{2}$,
  Huihua Yang$^{1,\dagger}$ \\[2pt]
  $^{1}$Beijing University of Posts and Telecommunications,
  $^{2}$Soochow University \\[2pt]
  \texttt{Jinzhe@bupt.edu.cn} \hspace{0.3cm}
  \texttt{linzhimin327@gmail.com} \hspace{0.3cm}
  \texttt{yhh@bupt.edu.cn} \\[2pt]
  \faGithub\ \url{https://github.com/jinzz831/DAGC}
}
}

\iclrfinalcopy 
\begin{document}

\maketitle
\lhead{}
\let\thefootnote\relax\footnotetext{$\ast$\; Equal contribution.}
\let\thefootnote\relax\footnotetext{$\dagger$\; Corresponding author.}

\begin{abstract}
Graph-based retrieval-augmented generation (RAG) provides a scalable paradigm for long-video understanding, but existing systems typically inherit a fixed temporal granularity from video segmentation when constructing their retrieval index. We argue that this design unnecessarily couples indexing granularity with evidence granularity: coarse representations can often suffice for locating relevant temporal regions, while fine-grained evidence remains important for downstream reasoning. We propose \textbf{Density-Aware Graph Construction (DAGC)}, a training-free approach that decouples a query-independent coarse retrieval index from the original fine-grained evidence space. DAGC constructs a compact, density-adaptive graph index by merging visually redundant neighboring chunks, while preserving mappings to the original temporal units. Retrieved coarse regions are subsequently expanded back to the original chunk granularity for fine-grained evidence refinement and answer generation. Experiments on MLVU, VideoMME, and LongVideoBench show that DAGC retains only about 40--50\% of the original graph nodes and achieves $1.3$--$1.7\times$ end-to-end wall-clock acceleration while preserving approximately 99\% of the original QA performance. The gains transfer across different LVLM backbones and video RAG pipelines, suggesting that long-video RAG need not maintain the same temporal granularity for indexing and evidence reasoning.
\end{abstract}

\input{sections/01_Introduction}

\input{sections/02_Related_Work}

\input{sections/03_Methodology}

\input{sections/04_Experiments}

\input{sections/05_Analysis_and_Discussion}
\input{sections/06_Conclusion}

\bibliography{iclr2027_conference}
\bibliographystyle{iclr2027_conference}

\clearpage
\appendix

\input{sections/07_Appendix}

\end{document}

%% file: math_commands.tex
\usepackage{amsmath,amsfonts,bm}

\def\eqref#1{equation~\ref{#1}}

\def\1{\bm{1}}

\DeclareMathAlphabet{\mathsfit}{\encodingdefault}{\sfdefault}{m}{sl}
\SetMathAlphabet{\mathsfit}{bold}{\encodingdefault}{\sfdefault}{bx}{n}



%% file: sections/01_Introduction.tex
\section{Introduction}

Long video understanding requires models to reason over extended temporal contexts and integrate information distributed across thousands of frames. Existing vision-language models (VLMs) typically address the resulting computational burden through sparse frame sampling or visual-token compression~\cite{song2024moviechatdensetokensparse, ren2024timechattimesensitivemultimodallarge, wang2025retakereducingtemporalknowledge}. However, these approaches face an inherent trade-off between efficiency and temporal fidelity: aggressive compression may discard important evidence, whereas dense representations incur substantial computational cost.

\begin{figure*}[t]
  \centering
  \includegraphics[width=\textwidth]{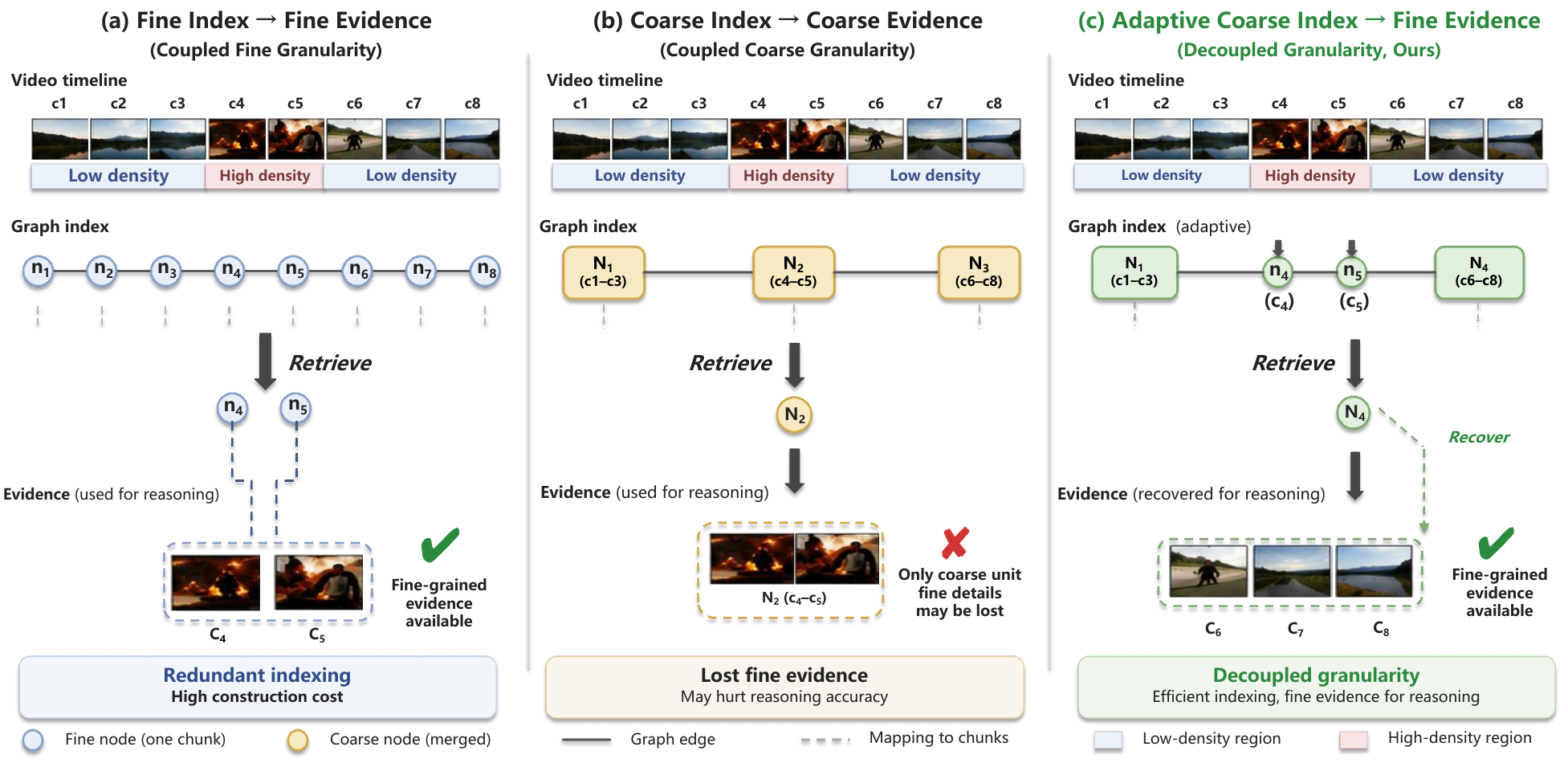}
  \caption{
  Decoupling indexing and evidence granularity in long-video RAG.
  Existing designs either maintain fine temporal resolution throughout the pipeline
  or coarsen both indexing and evidence.
  DAGC instead uses a density-adaptive coarse index for retrieval while preserving
  fine-grained evidence for downstream reasoning.
  }
  \label{fig:motivation}
\end{figure*}

Retrieval-Augmented Generation (RAG) offers a scalable alternative by retrieving relevant video regions before downstream reasoning~\cite{luo2025videoragvisuallyalignedretrievalaugmentedlong, jeong2025videoragretrievalaugmentedgenerationvideo}. Recent graph-based video RAG methods~\cite{shen2025vgentgraphbasedretrievalreasoningaugmentedgeneration, xu2025evragenhancinglongvideo} further organize video segments into structured graphs to capture temporal and semantic relationships. However, their graph granularity is typically inherited directly from fixed-length video segmentation, with each chunk represented as an individual node. This imposes a uniformly fine indexing resolution even though retrieval mainly requires locating relevant temporal regions, whereas downstream reasoning benefits from fine-grained visual evidence. As illustrated in Figure~\ref{fig:motivation}, \textbf{indexing and evidence reasoning therefore need not operate at the same temporal granularity}.

Based on this observation, we propose \textbf{Density-Aware Graph Construction (DAGC)}, a training-free approach that decouples a query-independent coarse indexing representation from the fine-grained evidence space used for downstream reasoning in long-video graph RAG. Long videos exhibit highly non-uniform temporal redundancy: rapidly changing regions require fine-grained representation, while visually stable regions can often be indexed more coarsely. DAGC therefore adaptively merges neighboring chunks according to adjacent visual similarity, subject to a maximum merging window $W$, and constructs the graph over the resulting coarse units. Each coarse node retains the indices of its constituent original chunks. Retrieval is performed on the query-independent compact graph index, after which selected regions are mapped back to the original chunks for fine-grained refinement and answer generation. This design reduces computation in the query-independent indexing stage, while preserving the original fine-grained evidence for post-retrieval reasoning.

We evaluate DAGC on MLVU, VideoMME, and LongVideoBench. Across different LVLM backbones and video RAG pipelines, DAGC retains approximately 40--50\% of the original indexing units and achieves $1.3$--$1.7\times$ end-to-end wall-clock acceleration while preserving about 99\% of the original QA performance. These results demonstrate that efficient long-video RAG does not require a uniformly fine representation throughout the pipeline: coarse indexing can eliminate substantial redundancy while fine-grained evidence remains available when needed for downstream reasoning.

%% file: sections/02_Related_Work.tex
\section{Related Work}

\subsection{Long Video Understanding with VLMs}

Long video understanding remains challenging for vision-language models
(VLMs) due to the large number of visual tokens required to represent
extended temporal contexts~\cite{song2024moviechatdensetokensparse,
ren2024timechattimesensitivemultimodallarge}. Existing approaches mainly
address this challenge by reducing the visual representation burden or
improving temporal memory.

Token compression methods reduce visual redundancy before or during
multimodal inference. Representative works~\cite{wang2025retakereducingtemporalknowledge,jiang2025stormtokenefficientlongvideo,liu2025keyframeorientedvisiontokenpruning} select informative frames or compress token sequences to fit long videos into limited context windows. Other approaches, such as VideoTree~\cite{wang2025videotreeadaptivetreebasedvideo}, organize video representations hierarchically and dynamically allocate representation capacity according to query requirements. However, these methods mainly optimize the amount or resolution of visual information within the representation used for downstream inference, and therefore still face a trade-off between reducing computation and preserving fine-grained temporal evidence.

Graph-based memory approaches provide an alternative by organizing video
content into structured representations of entities, events, and temporal
relationships~\cite{chu2025understandinglongvideosllmpowered}. While such
structures improve long-range reasoning, the temporal granularity of the
graph index is typically determined by a predefined video segmentation
scheme. This leaves largely unexplored whether the representation used for
indexing must retain the same temporal resolution as the evidence required
for downstream reasoning. Our work studies this complementary design
dimension by decoupling the two: the graph index can operate at an
adaptively coarser temporal granularity, while access to the original
fine-grained video evidence is preserved for post-retrieval reasoning.

\subsection{Retrieval-Augmented Generation for Video}

Retrieval-Augmented Generation (RAG) provides a scalable solution for long
video understanding by retrieving relevant temporal regions before
multimodal reasoning, avoiding the need to process the entire video at once
~\cite{lewis2021retrievalaugmentedgenerationknowledgeintensivenlp}.
Early video RAG approaches retrieve relevant clips or frames through dense
similarity matching over visual and textual representations~\cite{luo2025videoragvisuallyalignedretrievalaugmentedlong,
jeong2025videoragretrievalaugmentedgenerationvideo}. More recent methods
introduce structured representations to improve retrieval over long temporal contexts. For example, VideoRAG~\cite{ren2025videoragretrievalaugmentedgenerationextreme}
combines graph-based textual knowledge with multimodal visual retrieval for
extreme-length videos, while E-VRAG~\cite{xu2025evragenhancinglongvideo}
reduces retrieval computation through lightweight VLM scoring and similarity filtering.

Among graph-based video RAG methods, Vgent~\cite{shen2025vgentgraphbasedretrievalreasoningaugmentedgeneration} constructs semantic video graphs where clips are connected through shared entities and introduces structured reasoning to refine retrieved evidence. However, its graph index is constructed over fixed-length temporal chunks, such that the indexing granularity is directly inherited from the underlying video segmentation. Existing video RAG research has largely focused on improving which indexed units are retrieved or how retrieval is performed, while the granularity at which those units should be indexed has received less attention. Our work studies this complementary question by decoupling graph indexing from downstream evidence resolution: DAGC constructs a compact, density-adaptive coarse graph index while preserving mappings to the original chunks for fine-grained evidence recovery after retrieval.

\subsection{Video Segmentation and Scene Detection}

Video segmentation aims to divide videos into temporally coherent units and
has been widely studied for video understanding. Early approaches detect shot boundaries based on pixel-level differences, while supervised methods such as LGSS~\cite{rao2020localtoglobalapproachmultimodalmovie} learn scene
boundaries using multimodal features. More recently, MDLSeg
~\cite{mahon2025parameterfreevideosegmentationvision} formulates video
segmentation as an optimization problem based on the minimum description
length (MDL) principle, determining boundaries without manually specified
thresholds and improving downstream long-video understanding tasks.

Although video segmentation and DAGC both adapt temporal granularity, they
optimize it for different purposes. Scene segmentation seeks a temporally
coherent partition that reflects semantic or narrative boundaries. DAGC,
instead, treats temporal granularity as a retrieval-system design variable:
its goal is not to discover a single semantically correct partition, but to
construct an indexing representation that can be coarsened where temporal
redundancy permits. Importantly, this coarser indexing granularity does not
replace the original temporal units, which remain available for fine-grained evidence reasoning after retrieval.

\subsection{Graph Construction for RAG}

Graph-based RAG has become an effective paradigm for organizing structured
knowledge and improving retrieval quality in language applications.
GraphRAG~\cite{edge2025localglobalgraphrag} organizes entities and relations into semantic communities for global retrieval, while LightRAG
~\cite{guo2025lightragsimplefastretrievalaugmented} introduces efficient
dual-level indexing for scalable graph retrieval. NodeRAG ~\cite{xu2025noderagstructuringgraphbasedrag} further explores heterogeneous node structures to improve retrieval efficiency. Recent studies also investigate reducing graph construction cost by replacing expensive LLM-based extraction with lightweight alternatives
~\cite{min2025practicalgraphragefficientknowledge}.

Different from these efforts that primarily optimize graph extraction,
indexing structures, or retrieval strategies in text-based RAG, DAGC focuses on a complementary design dimension: the temporal granularity at which video content is represented in the graph index. Rather than requiring the graph index to retain the same fine temporal resolution used for downstream evidence reasoning, DAGC constructs a compact, density-adaptive coarse index while preserving access to the original temporal units. Retrieved regions are then mapped back to fine-grained evidence before downstream reasoning. This design reduces graph construction and indexing overhead without permanently coarsening the evidence available after retrieval.

%% file: sections/03_Methodology.tex
\begin{figure*}[t]
  \centering
  \includegraphics[width=\textwidth]{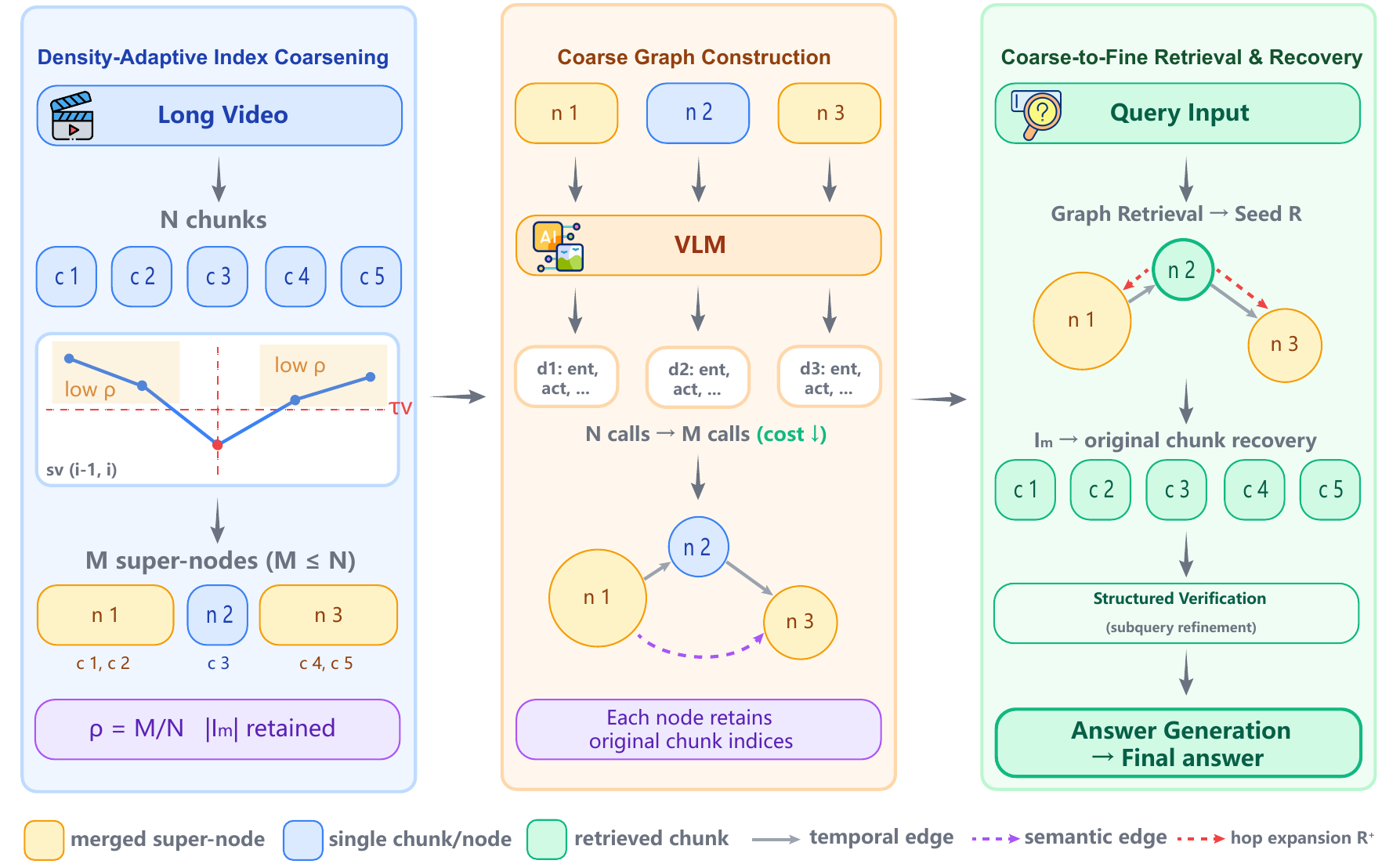}
  \caption{
  Overview of Density-Aware Graph Construction (DAGC).
  DAGC separates the temporal granularity used for graph indexing from that
  used for downstream evidence reasoning.
  Visually redundant neighboring chunks are merged into density-adaptive
  coarse nodes for efficient indexing, while each node preserves the mapping
  $\mathcal{I}_m$ to its constituent original chunks.
  Retrieval first operates on the compact graph for coarse localization and
  then recovers the original chunks for fine-grained refinement and answer
  generation.
  }
  \label{fig:pipeline}
\end{figure*}

\section{Density-Aware Graph Construction with Granularity Decoupling}
\label{sec:methodology}

Existing graph-based video RAG systems typically use the same temporal units
for two distinct purposes: constructing the retrieval index and providing
visual evidence for downstream reasoning.
However, these two stages impose different requirements.
Retrieval primarily needs an efficient representation for locating relevant
temporal regions, whereas final reasoning benefits from fine-grained access
to the original visual evidence.
We therefore propose \textbf{Density-Aware Graph Construction (DAGC)}, a
training-free coarse-to-fine design that explicitly decouples these two
granularities.

DAGC constructs a query-independent, density-adaptive coarse graph index
for candidate localization while preserving a direct mapping from every
coarse node to its constituent original chunks. Consequently, compression
is applied to the \emph{indexing representation} rather than permanently
to the evidence available for reasoning. As illustrated in
Figure~\ref{fig:pipeline}, DAGC implements this decoupling through three
stages: \textbf{(1)} density-adaptive index coarsening via neighboring-chunk
merging, \textbf{(2)} coarse graph index construction, and
\textbf{(3)} coarse-to-fine evidence recovery.

\subsection{Density-Adaptive Index Coarsening}
\label{sec:node_merging}

We divide the input video $X$ into $N$ fixed-length chunks
$\{c_1,c_2,\ldots,c_N\}$, each containing $K$ sampled frames.
Rather than assigning the same indexing resolution to every temporal region,
DAGC allocates graph granularity according to local temporal redundancy.
Importantly, the objective is not to recover a semantically complete event
partition, but to determine where multiple adjacent chunks can share a
coarser indexing representation without removing access to their original
evidence.

We use adjacent visual similarity as a lightweight proxy for this local
redundancy. For each chunk, frame-level visual features are temporally pooled
and $L_2$-normalized, and adjacent similarity is computed as
\begin{equation}
\mathbf{v}_i =
\operatorname{Norm}\!\left(\operatorname{Pool}(c_i)\right),
\qquad
s_v(i-1,i)=\mathbf{v}_{i-1}^{\top}\mathbf{v}_i .
\label{eq:visual_similarity}
\end{equation}

A high similarity score indicates that neighboring chunks carry redundant
visual content and can share a coarse indexing unit, whereas a low score
suggests a transition where finer indexing resolution should be preserved.
We greedily merge adjacent chunks when
\begin{equation}
s_v(i-1,i)\geq\tau_v
\quad\text{and}\quad
|\mathcal{I}_m|<W,
\label{eq:merge_rule}
\end{equation}
where $\tau_v$ is the visual-similarity threshold, $W$ bounds the maximum
merging span, and $\mathcal{I}_m$ records the original chunk indices assigned
to super-node $n_m$.
The span constraint prevents long visually stable regions from collapsing
into excessively coarse indexing units.

After adaptation, the $N$ original chunks are represented by $M\leq N$
coarse units.
For each merged unit $n_m$, frames from its constituent chunks are
concatenated in temporal order and uniformly re-sampled to the same
$K$-frame budget as an original chunk, producing $\tilde{c}_m$ for semantic
extraction.
Thus, increasing the temporal coverage of a coarse node does not increase
its per-node visual input budget.
Meanwhile, the original chunks themselves are retained through
$\mathcal{I}_m$ for subsequent fine-grained recovery.
We denote the retained indexing-unit ratio as $\rho=M/N$.

\subsection{Bounded-Cost Coarse Graph Construction}
\label{sec:compressed_graph}

The adapted units define a second, coarser temporal resolution used
specifically for graph indexing.
Conventional fixed-granularity construction performs semantic extraction
independently for all $N$ original chunks.
DAGC instead performs the expensive graph-construction stage only on the
$M$ coarse units.

For each $\tilde{c}_m$, the LVLM extracts the structured semantics required
by the underlying graph-RAG framework, including entities, actions, scenes,
and textual descriptions.
Because every coarse unit is restricted to the same $K$-frame budget,
reducing $N$ indexing units to $M$ directly reduces the number of
query-independent LVLM extraction calls rather than shifting computation
into larger per-node visual inputs.

In our primary Vgent
instantiation~\cite{shen2025vgentgraphbasedretrievalreasoningaugmentedgeneration},
semantically related entities are matched to global prototypes and nodes
sharing semantic information are connected to form the coarse retrieval
graph $\mathcal{G}_c$.
DAGC is designed to be orthogonal to this semantic graph definition:
it changes the temporal units on which graph semantics are instantiated,
allowing the same graph-RAG machinery to operate over a compact,
density-adaptive index.
Each graph node additionally stores $\mathcal{I}_m$, establishing the
connection between the coarse indexing space and the original temporal
evidence space.

\subsection{Coarse-to-Fine Evidence Retrieval}
\label{sec:retrieval_recovery}

DAGC deliberately assigns different representations to candidate localization
and final evidence reasoning.
The coarse graph is used to efficiently identify relevant temporal regions,
but its compressed nodes are not treated as the final visual evidence.
Instead, retrieval proceeds in two resolutions: coarse graph localization
followed by original-granularity evidence recovery.

Given a question $Q$, query-related information is first matched against
$\mathcal{G}_c$ to rank coarse candidate nodes.
In our Vgent instantiation, the corresponding graph retrieval procedure is
used to obtain the highest-ranked seed set $\mathcal{R}_s$.
The seed regions are subsequently expanded to temporally related candidates,
yielding $\mathcal{R}_s^{+}$.

The retrieved coarse regions are then projected back to the original temporal
units through their stored mappings:
\begin{equation}
\mathcal{C}_{\mathrm{cand}}
=
\bigcup_{n_m\in\mathcal{R}_s^{+}}
\left\{c_i \mid i\in\mathcal{I}_m\right\}.
\label{eq:evidence_recovery}
\end{equation}

The resulting $\mathcal{C}_{\mathrm{cand}}$ contains original-granularity
chunks rather than compressed super-nodes.
These chunks are reranked and verified using question-specific refinement,
and the selected fine-grained visual evidence, together with intermediate
reasoning results, is finally provided to the LVLM for answer generation.

This coarse-to-fine retrieval design is the key distinction between DAGC
and conventional representation compression.
DAGC compresses the representation used to \emph{search} the video, while
preserving the finer representation used to \emph{reason} about retrieved
evidence.
It therefore reduces redundant graph construction and indexing computation
without forcing downstream reasoning to operate at the same coarse temporal
resolution.

%% file: sections/04_Experiments.tex
\section{Experiments}
\label{sec:experiments}

\subsection{Experimental Settings}
\label{sec:experimental_settings}

\paragraph{Baselines.}
We primarily instantiate DAGC on Vgent~\cite{shen2025vgentgraphbasedretrievalreasoningaugmentedgeneration},
a graph-based retrieval-reasoning framework for long-video understanding.
We evaluate three Qwen-family LVLM backbones:
Qwen2.5-VL-7B, Qwen2.5-VL-3B, and Qwen2-VL-7B.
We compare against the corresponding vanilla LVLMs, which directly perform
inference on uniformly sampled video frames, and Vgent, which constructs a
fixed-granularity video graph.
To evaluate transferability, we additionally apply DAGC to
InternVL3.5-8B with the Vgent pipeline and to VideoRAG as a different
long-video RAG framework.

\paragraph{Benchmarks.}
We evaluate on MLVU~\cite{zhou2025mlvubenchmarkingmultitasklong},
VideoMME~\cite{fu2025videommefirstevercomprehensiveevaluation}, and
LongVideoBench (LVB)~\cite{wu2024longvideobenchbenchmarklongcontextinterleaved}.
Together, these benchmarks cover diverse long-video understanding tasks,
including counting, temporal ordering, visual grounding, topic reasoning,
and fine-grained referred video understanding.

\paragraph{Implementation Details.}
All experiments are conducted on NVIDIA A100 40GB GPUs.
Following Vgent, videos are sampled at 1 FPS and divided into 64-frame chunks.
Unless otherwise specified, DAGC uses a visual-similarity threshold
$\tau_v=0.95$, a maximum merging window $W=3$, and a fixed 64-frame input
budget for each merged super-node.
During retrieval, we retain $k_s=12$ coarse seeds before temporal expansion
and original-chunk recovery.
The same DAGC configuration is used across benchmarks and LVLM backbones,
while other retrieval and refinement settings follow the corresponding
RAG backbone.

\paragraph{Metrics.}
We report multiple-choice accuracy for effectiveness and the retained
indexing-unit ratio (\textbf{Retained}) for compression.
Efficiency is measured by query-independent \textbf{Offline Speedup} and
end-to-end \textbf{Wall Speedup}.
For normalized runtime analysis, we additionally report offline, online,
and first-query time in seconds per minute of video.

\subsection{Main Results}
\label{subsec:main_results}

\begin{table*}[t]
\centering
\caption{
Main results on three long-video benchmarks.
\textbf{Accuracy} denotes multiple-choice accuracy,
\textbf{Retained} denotes the retained graph-node ratio relative to Vgent,
and \textbf{Wall Speedup} denotes end-to-end wall-clock acceleration
relative to the corresponding Vgent baseline.
\textbf{Performance Retention} is computed from the average accuracy across
the three benchmarks.
}
\setlength{\tabcolsep}{2.6pt}
\renewcommand{\arraystretch}{1.02}
\resizebox{\textwidth}{!}
{
\begin{tabular}{lccccccccccc}
\toprule
\multirow{2}{*}{\textbf{Model}} &
\multicolumn{3}{c}{\textbf{MLVU}} &
\multicolumn{3}{c}{\textbf{VideoMME}} &
\multicolumn{3}{c}{\textbf{LVB}} &
\multirow{2}{*}{\shortstack{\textbf{Avg.}\\\textbf{Accuracy}}} &
\multirow{2}{*}{\shortstack{\textbf{Performance}\\\textbf{Retention}}} \\
\cmidrule(lr){2-4}
\cmidrule(lr){5-7}
\cmidrule(lr){8-10}
&
\textbf{Accuracy} &
\textbf{Retained} &
\shortstack{\textbf{Wall}\\\textbf{Speedup}} &
\textbf{Accuracy} &
\textbf{Retained} &
\shortstack{\textbf{Wall}\\\textbf{Speedup}} &
\textbf{Accuracy} &
\textbf{Retained} &
\shortstack{\textbf{Wall}\\\textbf{Speedup}} &
&
\\
\midrule

Qwen2.5-VL-7B
& 69.0 & -- & --
& 70.1 & -- & --
& 59.4 & -- & --
& 66.2 & -- \\

 + Vgent
& 73.3 & 100\% & 1.0$\times$
& 73.3 & 100\% & 1.0$\times$
& 63.3 & 100\% & 1.0$\times$
& 70.0 & 100\% \\

\rowcolor{oursbg}
+ DAGC
& 73.4 & 45\% & 1.3$\times$
& 71.1 & 45\% & 1.4$\times$
& 63.1 & 47\% & 1.6$\times$
& 69.2 & \textcolor{retcolor}{99\%} \\

\midrule

Qwen2.5-VL-3B
& 65.0 & -- & --
& 67.0 & -- & --
& 56.3 & -- & --
& 62.8 & -- \\

 + Vgent
& 70.0 & 100\% & 1.0$\times$
& 69.0 & 100\% & 1.0$\times$
& 60.0 & 100\% & 1.0$\times$
& 66.3 & 100\% \\

\rowcolor{oursbg}
 + DAGC
& 69.9 & 45\% & 1.3$\times$
& 66.2 & 45\% & 1.3$\times$
& 60.6 & 47\% & 1.3$\times$
& 65.6 & \textcolor{retcolor}{99\%} \\

\midrule

Qwen2-VL-7B
& 65.7 & -- & --
& 68.6 & -- & --
& 56.1 & -- & --
& 63.5 & -- \\

+ Vgent
& 71.7 & 100\% & 1.0$\times$
& 69.7 & 100\% & 1.0$\times$
& 58.9 & 100\% & 1.0$\times$
& 66.8 & 100\% \\

\rowcolor{oursbg}
+ DAGC
& 72.1 & 45\% & 1.5$\times$
& 67.3 & 45\% & 1.4$\times$
& 58.6 & 47\% & 1.7$\times$
& 66.0 & \textcolor{retcolor}{99\%} \\

\bottomrule
\end{tabular}
}

\label{tab:main_results}
\end{table*}

Table~\ref{tab:main_results} summarizes the effectiveness--efficiency
trade-off of DAGC across three benchmarks and three Qwen-family backbones.
DAGC retains only 45\%--47\% of the original Vgent graph nodes and achieves
1.3$\times$--1.7$\times$ end-to-end wall-clock acceleration.
Despite removing more than half of the indexing units, the average accuracy
decreases by only 0.7--0.8 percentage points across the three backbones,
corresponding to approximately 99\% performance retention.

Although the accuracy changes vary across individual benchmarks, the overall
performance remains largely preserved under substantial graph compression.
These results indicate that fixed-granularity graph construction contains
considerable temporal redundancy, and that a compact coarse index can support
efficient retrieval while fine-grained evidence is recovered for downstream
reasoning.



\subsection{Efficiency Analysis}
\label{subsec:efficiency_analysis}
\begin{table}[t]
\centering
\caption{
Normalized runtime comparison using Qwen2.5-VL-7B.
\textbf{Offline Time} denotes query-independent graph construction,
\textbf{Online Time} denotes query-dependent inference after graph construction,
and \textbf{First-query Time} includes both stages.
All values are reported in seconds per minute of video.
}
\small
\setlength{\tabcolsep}{4pt}
\renewcommand{\arraystretch}{1.05}

\begin{tabular}{lccc}
\toprule
\textbf{Model} &
\textbf{Offline Time} &
\textbf{Online Time} &
\textbf{First-query Time} \\
\midrule

Qwen2.5-VL-7B
& --
& --
& 3.12 \\

Qwen2.5-VL-7B + Vgent
& 24.14
& 4.12
& 28.26 \\

\rowcolor{oursbg}
Qwen2.5-VL-7B + DAGC
& \textbf{10.73}
& 4.24
& \textbf{14.97} \\

\bottomrule
\end{tabular}

\label{tab:inference_time}
\end{table}
Table~\ref{tab:inference_time} shows that DAGC substantially reduces
query-independent graph-construction cost.
Offline time decreases from 24.14 to 10.73 seconds per minute of video,
a 55.6\% reduction, while online inference increases only slightly
from 4.12 to 4.24 seconds due to coarse-to-fine evidence recovery.
Consequently, first-query time is reduced from 28.26 to 14.97 seconds
per minute of video (47.0\%).
These results show that the offline savings introduced by coarse indexing substantially outweigh the small online recovery overhead. This cost decomposition highlights a key distinction of DAGC: the computational savings arise primarily from the query-independent indexing stage, allowing the reduced graph-construction cost to be amortized across subsequent queries while preserving fine-grained evidence for online reasoning.

\subsection{Generalization Across Models and RAG Frameworks}
\label{subsec:generalization}

The previous experiments evaluate DAGC under the Qwen--Vgent configuration.
We next examine whether its compression behavior transfers across both
LVLM families and video-RAG pipelines.

\begin{table}[t]
\centering
\caption{
Cross-model-family evaluation using InternVL3.5-8B.
\textbf{Retained} denotes the retained graph-node ratio relative to Vgent,
and \textbf{Wall Speedup} is measured relative to the corresponding
InternVL3.5-8B + Vgent baseline.
}
\small
\setlength{\tabcolsep}{4pt}
\renewcommand{\arraystretch}{1.05}
{
\begin{tabular}{lccccc}
\toprule
\textbf{Dataset} &
\shortstack{\textbf{Vgent}\\\textbf{Accuracy}} &
\shortstack{\textbf{DAGC}\\\textbf{Accuracy}} &
$\boldsymbol{\Delta}$\textbf{ Accuracy} &
\textbf{Retained} &
\shortstack{\textbf{Wall}\\\textbf{Speedup}} \\
\midrule

LVB
& 63.07
& 63.05
& $-0.02$
& 47\%
& 1.3$\times$ \\

MLVU
& 73.46
& 73.04
& $-0.42$
& 47\%
& 1.4$\times$ \\

VideoMME
& 68.18
& 65.63
& $-2.55$
& 47\%
& 1.4$\times$ \\

\bottomrule
\end{tabular}
}

\label{tab:runtime}
\end{table}




\begin{table*}[t]
\centering
\caption{
Cross-framework evaluation after integrating DAGC into VideoRAG.
\textbf{Common N} denotes the number of examples successfully evaluated
by both the baseline and DAGC.
\textbf{Retained} denotes the retained temporal indexing-unit ratio relative
to the uncompressed VideoRAG baseline.
\textbf{Offline Speedup} measures query-independent index construction,
while \textbf{Wall Speedup} measures complete experimental wall-clock
acceleration.
}
\small
\setlength{\tabcolsep}{4pt}
\renewcommand{\arraystretch}{1.05}

\begin{tabular*}{\textwidth}{
@{\extracolsep{\fill}}
l
c
c
c
c
c
c
c
@{}
}
\toprule

\textbf{Dataset} &
\textbf{Common N} &
\shortstack{\textbf{Baseline}\\\textbf{Accuracy}} &
\shortstack{\textbf{DAGC}\\\textbf{Accuracy}} &
$\boldsymbol{\Delta}$\textbf{ Accuracy} &
\textbf{Retained} &
\shortstack{\textbf{Offline}\\\textbf{Speedup}} &
\shortstack{\textbf{Wall}\\\textbf{Speedup}} \\

\midrule

LVB
& 1,296
& 52.70\%
& \textbf{54.17\%}
& \textbf{+1.47 pp}
& 48.34\%
& 1.377$\times$
& 1.309$\times$ \\

MLVU
& 2,174
& \textbf{64.49\%}
& 62.88\%
& $-1.61$ pp
& 40.39\%
& \textbf{1.695$\times$}
& \textbf{1.328$\times$} \\

VideoMME
& 2,683
& 65.97\%
& \textbf{66.72\%}
& \textbf{+0.75 pp}
& 45.10\%
& 1.455$\times$
& 1.308$\times$ \\

\bottomrule
\end{tabular*}

\label{tab:videorag_generalization}
\end{table*}

Table~\ref{tab:videorag_generalization} shows that DAGC retains only
40.39\%--48.34\% of VideoRAG indexing units, achieving
1.377$\times$--1.695$\times$ offline speedup and over
1.30$\times$ wall-clock acceleration across all three benchmarks.
Accuracy improves on LVB and VideoMME but decreases on MLVU, indicating
that the cross-framework benefit of DAGC is primarily computational rather
than a consistent accuracy improvement.

Together with the InternVL3.5-8B results, these experiments show that
DAGC's efficiency benefit is not limited to a single LVLM family or RAG
pipeline.

%% file: sections/05_Analysis_and_Discussion.tex
\section{Analysis and Discussion}
\label{sec:analysis_discussion}

We further examine three questions underlying the design of DAGC: whether indexing and downstream evidence reasoning require the same temporal granularity, whether indexing granularity should adapt to local temporal redundancy, and whether semantic event boundaries provide a suitable alternative basis for temporal partitioning. Together, these analyses help disentangle the roles of granularity decoupling, density-aware adaptation, and temporal partition choice in the effectiveness of DAGC.

\subsection{Indexing and Evidence Granularity}
\label{subsec:component_ablation}

A key design principle of DAGC is to decouple the temporal granularity
used for indexing from that used for downstream evidence reasoning.
To verify this design, we compare DAGC variants with and without
coarse indexing and fine-grained evidence recovery.

\begin{table}[t]
\centering
\caption{
Ablation of coarse indexing and fine-grained evidence recovery on
LongVideoBench using Qwen2.5-VL-7B.
SN denotes super-node compression, RR denotes recovery and reranking
of original chunks, and TE denotes temporal expansion.
Detailed category-level results are reported in Appendix~\ref{app:component_ablation}.
}
\small
\setlength{\tabcolsep}{4pt}
\renewcommand{\arraystretch}{1.05}

\begin{tabular}{lccc}
\toprule
\textbf{Variant} &
\textbf{Retained} &
\textbf{Accuracy} &
\textbf{Wall Speedup} \\
\midrule

Qwen2.5-VL-7B
& --
& 59.4
& -- \\

+ Vgent
& 100\%
& 63.3
& 1.0$\times$ \\

+ DAGC w/o SN
& 100\%
& 62.6
& 1.0$\times$ \\

+ DAGC w/o RR
& 47\%
& 60.6
& 1.6$\times$ \\

+ DAGC w/o TE
& 47\%
& 62.2
& 1.6$\times$ \\

\rowcolor{oursbg}
+ DAGC
& 47\%
& \textbf{63.1}
& 1.6$\times$ \\

\bottomrule
\end{tabular}

\label{tab:component_ablation}
\end{table}

Table~\ref{tab:component_ablation} examines the contribution of
coarse indexing and fine-grained evidence recovery.
Directly using compressed super-nodes as downstream evidence
(\textit{w/o RR}) reduces accuracy to 60.6, indicating that coarse
representations alone are insufficient for final reasoning.
Recovering and reranking the original chunks substantially restores
performance, while temporal expansion further improves retrieval
completeness.

With the complete coarse-to-fine pipeline, DAGC achieves 63.1 accuracy,
close to the 63.3 achieved by the original fine-grained Vgent baseline,
while retaining only 47\% of graph nodes.
These results support the central hypothesis of DAGC:
the retrieval index can operate at a coarser temporal granularity,
while fine-grained evidence can be recovered after retrieval for
accurate downstream reasoning.

\subsection{Density-aware Compression Strategy}
\label{subsec:compression_strategy}

Although DAGC reduces the number of graph nodes, the improvement should
not come merely from node reduction. We therefore compare DAGC with
content-agnostic compression strategies under the same retained-node
budget.

\begin{table}[t]
\centering
\caption{
Comparison of compression strategies on LongVideoBench using
Qwen2.5-VL-7B under the same retained-node budget.
}
\small
\setlength{\tabcolsep}{4pt}
\renewcommand{\arraystretch}{1.05}

\begin{tabular}{lcc}
\toprule
\textbf{Method} &
\textbf{Accuracy} &
\textbf{Retained} \\
\midrule

Random Merge
& 61.53
& 47\% \\

Uniform Merge
& 62.28
& 47\% \\

\rowcolor{oursbg}
DAGC
& \textbf{63.07}
& 47\% \\

\bottomrule
\end{tabular}

\label{tab:compression_strategy}
\end{table}










Table~\ref{tab:compression_strategy} compares different node selection
strategies with an identical 47\% retained-node ratio.
DAGC improves over Uniform Merge and Random Merge by 0.79 and
1.54 percentage points, respectively.

By preserving fine-grained representations in regions with larger
temporal variation, DAGC achieves compression while maintaining
downstream reasoning capability.

\subsection{Event Boundary Analysis}
\label{subsec:event_boundary_analysis}

An alternative approach to adaptive temporal granularity is to rely on
explicit event segmentation. To examine whether semantic event boundaries
provide a better merging criterion, we incorporate EfficientGEBD
boundaries as hard constraints that prevent merging across predicted
event transitions, while keeping the downstream retrieval and reasoning
pipeline unchanged.

\begin{table}[!htbp]
\centering
\caption{
Effect of EfficientGEBD event-boundary constraints on the complete
Order, Needle, and Count subsets (820 questions).
$\Delta$ denotes percentage-point change relative to DAGC.
}
\small
\setlength{\tabcolsep}{4pt}
\renewcommand{\arraystretch}{1.05}
\begin{tabular}{lccc}
\toprule
\textbf{Task} &
\textbf{DAGC} &
\textbf{+ Event Boundary} &
$\boldsymbol{\Delta}$ \\
\midrule
Order  & 70.66 & \textbf{72.97} & +2.32 \\
Needle & \textbf{82.25} & 81.69 & -0.56 \\
Count  & \textbf{60.68} & 57.77 & -2.91 \\
\midrule
Weighted Overall & \textbf{73.17} & 72.93 & -0.24 \\
\bottomrule
\end{tabular}
\label{tab:event_boundary}
\end{table}

As shown in Table~\ref{tab:event_boundary}, explicit event boundaries
improve temporal ordering performance but degrade Needle and Count
accuracy, resulting in a small overall decrease of 0.24 percentage
points.

This indicates that perceptual event transitions are not always aligned
with the evidence granularity required by long-video question answering.
Therefore, DAGC is designed as a redundancy-aware indexing strategy
rather than a semantic video segmentation method.
Additional boundary-aware experiments are reported in
Appendix~\ref{app:event_boundary_analysis}.

%% file: sections/06_Conclusion.tex
\section{Conclusion}

In this paper, we present Density-Aware Graph Construction (DAGC), a training-free approach that decouples indexing granularity from evidence granularity for efficient long-video graph RAG. DAGC constructs a compact, density-adaptive coarse graph index by merging visually redundant neighboring chunks, while preserving mappings to the original temporal units for fine-grained evidence recovery after retrieval. Across three long-video benchmarks, DAGC retains approximately 40--50\% of the original graph nodes and achieves $1.3$--$1.7\times$ end-to-end wall-clock acceleration while preserving about 99\% of the original QA performance. Experiments across different LVLM backbones and video RAG pipelines further demonstrate that this design transfers beyond a single model or framework. 
Our analysis also shows that explicit event boundaries do not consistently improve downstream performance, suggesting that long-video RAG need not rely on a single semantic partition or temporal granularity throughout the pipeline. 
Instead, coarse indexing can be combined with fine-grained evidence recovery to reduce redundant computation while preserving access to detailed visual evidence.

%% file: sections/07_Appendix.tex
\clearpage
\onecolumn

\section{Experimental Results}

\newcommand{\tablesize}{\fontsize{8.2pt}{10pt}\selectfont}
\renewcommand{\arraystretch}{1.12}


\subsection{Detailed Results on MLVU}
\label{app:mlvu_detailed}

Table~\ref{tab:main_results_on_MLVU} reports the category-level
performance on MLVU across the seven multiple-choice tasks.
Overall, DAGC largely preserves the performance of the original
Vgent pipeline after graph compression. With Qwen2.5-VL-7B,
DAGC achieves an overall accuracy of 73.4 compared with 73.3 for
Vgent, while Qwen2-VL-7B improves from 71.7 to 72.1.
For Qwen2.5-VL-3B, the difference is only 0.1 percentage points.
At the task level, the effect of compression varies across
categories, suggesting that redundant graph nodes can be removed
without systematically degrading the different reasoning abilities
evaluated by MLVU.

\begin{table*}[h]
\centering
\caption{
Detailed results on MLVU across seven tasks.
Count, Ego, Needle, Order, PlotQA, Topic, and Anomaly are the
seven evaluated tasks. Overall denotes the aggregate accuracy
across all seven tasks.
}
\tablesize
\setlength{\tabcolsep}{2.4pt}

\begin{tabular}{lcccccccc}
\toprule
\textbf{Model} &
\textbf{Count} &
\textbf{Ego} &
\textbf{Needle} &
\textbf{Order} &
\textbf{PlotQA} &
\textbf{Topic} &
\textbf{Anomaly} &
\textbf{Overall} \\
\midrule

Qwen2.5-VL-7B
& 42.3 & 60.0 & 79.4 & 66.7 & 75.1 & 86.4 & 73.0 & 69.0 \\

Qwen2.5-VL-7B + Vgent
& 59.6 & 61.4 & 81.1 & 73.4 & 76.1 & 87.1 & 74.5 & 73.3 \\

\rowcolor{oursbg}
Qwen2.5-VL-7B + DAGC
& 60.7 & 62.0 & 82.3 & 70.7 & 76.6 & 87.1 & 74.0 & 73.4 \\

\midrule

Qwen2.5-VL-3B
& 32.3 & 52.9 & 78.2 & 56.2 & 71.5 & 88.1 & 76.0 & 65.0 \\

Qwen2.5-VL-3B + Vgent
& 53.3 & 58.0 & 80.0 & 62.5 & 71.9 & 89.0 & 75.5 & 70.0 \\

\rowcolor{oursbg}
Qwen2.5-VL-3B + DAGC
& 52.4 & 58.2 & 78.6 & 64.0 & 72.1 & 88.3 & 75.5 & 69.9 \\

\midrule

Qwen2-VL-7B
& 33.2 & 66.1 & 79.4 & 53.6 & 71.1 & 86.6 & 70.2 & 65.7 \\

Qwen2-VL-7B + Vgent
& 61.2 & 67.7 & 82.2 & 61.0 & 71.5 & 87.1 & 71.0 & 71.7 \\

\rowcolor{oursbg}
Qwen2-VL-7B + DAGC
& 63.9 & 67.6 & 81.9 & 61.7 & 71.3 & 87.1 & 71.0 & 72.1 \\

\bottomrule
\end{tabular}

\label{tab:main_results_on_MLVU}
\end{table*}


\subsection{Detailed Results on VideoMME}
\label{app:videomme_detailed}

Table~\ref{tab:main_results_on_videomme} further breaks down
the VideoMME results according to video duration. DAGC retains
performance relatively well on short and medium videos, whereas
the long-video subset is more sensitive to graph compression.
This trend is consistent across the three evaluated backbones.
In particular, aggressive compression of long videos can merge
temporally extended regions in which sparse but important evidence
is distributed across multiple chunks. These results indicate that
the optimal compression strength can depend on video duration and
information density.

\begin{table*}[h]
\centering
\caption{
Detailed results on VideoMME across different video durations.
Short, Medium, and Long denote the three duration subsets.
Overall denotes the aggregate accuracy over all duration subsets.
Wall Speedup denotes the relative wall-clock speedup over the
corresponding Vgent baseline.
}
\tablesize
\setlength{\tabcolsep}{2.4pt}

\begin{tabular}{lccccc}
\toprule
\textbf{Model} &
\textbf{Short} &
\textbf{Medium} &
\textbf{Long} &
\shortstack{\textbf{Overall}\\\textbf{Accuracy}} &
\shortstack{\textbf{Wall}\\\textbf{Speedup}} \\
\midrule

Qwen2.5-VL-7B + Vgent
& 78.76 & 72.63 & 68.33 & 73.25 & 1.0$\times$ \\

\rowcolor{oursbg}
Qwen2.5-VL-7B + DAGC
& 78.67 & 72.41 & 62.22 & 71.11 & 1.4$\times$ \\

\midrule

Qwen2.5-VL-3B + Vgent
& 75.33 & 68.55 & 63.66 & 69.00 & 1.0$\times$ \\

\rowcolor{oursbg}
Qwen2.5-VL-3B + DAGC
& 74.23 & 65.22 & 59.11 & 66.19 & 1.3$\times$ \\

\midrule

Qwen2-VL-7B + Vgent
& 76.43 & 69.55 & 63.22 & 69.74 & 1.0$\times$ \\

\rowcolor{oursbg}
Qwen2-VL-7B + DAGC
& 75.55 & 66.55 & 59.77 & 67.30 & 1.4$\times$ \\

\bottomrule
\end{tabular}

\label{tab:main_results_on_videomme}
\end{table*}


\clearpage
\subsection{Detailed Results on LongVideoBench}
\label{app:lvb_detailed}

Tables~\ref{tab:detailed_results_on_lvb_part1}
and~\ref{tab:detailed_results_on_lvb_part2} report the complete
category-level results on LongVideoBench. We split the categories
into two tables for readability. DAGC exhibits small
category-dependent fluctuations relative to Vgent while retaining
approximately half of the original graph nodes. Improvements can
be observed in several temporal and object-relation categories,
whereas other categories experience moderate degradation.
Together with the runtime results, this comparison illustrates the
accuracy--efficiency trade-off introduced by adaptive graph
compression.

\begin{table*}[h]
\centering
\caption{
Detailed category-level results on LongVideoBench, Part I.
The table reports the first group of question-category accuracies.
}
\tablesize
\setlength{\tabcolsep}{2.4pt}

\begin{tabular}{lccccccccc}
\toprule
\textbf{Model} &
\textbf{E2O} &
\textbf{SSS} &
\textbf{T2E} &
\textbf{S2O} &
\textbf{SAA} &
\textbf{TAA} &
\textbf{S2A} &
\textbf{SOS} &
\textbf{T2A} \\
\midrule

Qwen2.5-VL-7B + Vgent
& 75.4 & 46.4 & 72.3 & 66.7 & 61.1 & 59.8 & 72.1 & 65.4 & 70.4 \\

\rowcolor{oursbg}
Qwen2.5-VL-7B + DAGC
& 75.4 & 44.8 & 69.2 & 68.1 & 58.3 & 58.5 & 70.5 & 65.4 & 67.9 \\

\midrule

Qwen2.5-VL-3B + Vgent
& 64.6 & 47.4 & 64.6 & 62.5 & 52.8 & 52.4 & 62.5 & 65.4 & 63.0 \\

\rowcolor{oursbg}
Qwen2.5-VL-3B + DAGC
& 64.6 & 43.8 & 66.2 & 63.9 & 54.2 & 53.7 & 62.5 & 64.2 & 64.2 \\

\midrule

Qwen2-VL-7B + Vgent
& 70.8 & 46.4 & 66.2 & 56.9 & 56.9 & 52.4 & 71.6 & 61.7 & 56.8 \\

\rowcolor{oursbg}
Qwen2-VL-7B + DAGC
& 70.8 & 47.4 & 66.2 & 55.6 & 52.8 & 53.7 & 72.7 & 63.0 & 59.3 \\

\bottomrule
\end{tabular}

\label{tab:detailed_results_on_lvb_part1}
\end{table*}

\begin{table*}[h]
\centering
\caption{
Detailed category-level results on LongVideoBench, Part II. The table reports the question-category accuracies and includes the overall accuracy, speedup, and retained-node ratio.
}
\tablesize
\setlength{\tabcolsep}{2.4pt}

\begin{tabular}{lcccccccc|ccc}
\toprule
\textbf{Model} &
\textbf{S2E} &
\textbf{T3O} &
\textbf{T2O} &
\textbf{O3O} &
\textbf{O2E} &
\textbf{T3E} &
\textbf{TOS} &
\textbf{E3E} &
\shortstack{\textbf{Overall}\\\textbf{Accuracy}} &
\shortstack{\textbf{Wall}\\\textbf{Speedup}} &
\textbf{Retained} \\
\midrule

Qwen2.5-VL-7B + Vgent
& 76.3 & 59.5 & 64.1 & 60.6 & 69.3 & 49.3 & 40.0 & 68.1
& 63.33 & 1.0$\times$ & -- \\

\rowcolor{oursbg}
Qwen2.5-VL-7B + DAGC
& 74.2 & 56.8 & 69.2 & 68.2 & 68.2 & 49.3 & 42.7 & 67.0
& 63.07 & 1.6$\times$ & 47\% \\

\midrule

Qwen2.5-VL-3B + Vgent
& 77.4 & 62.2 & 55.1 & 54.5 & 67.8 & 54.8 & 44.0 & 67.0
& 60.00 & 1.0$\times$ & -- \\

\rowcolor{oursbg}
Qwen2.5-VL-3B + DAGC
& 75.3 & 59.5 & 59.0 & 62.1 & 70.5 & 52.1 & 42.7 & 68.1
& 60.6 & 1.3$\times$ & 47\% \\

\midrule

Qwen2-VL-7B + Vgent
& 68.8 & 66.2 & 57.7 & 62.1 & 62.5 & 49.3 & 32.0 & 62.8
& 58.88 & 1.0$\times$ & -- \\

\rowcolor{oursbg}
Qwen2-VL-7B + DAGC
& 71.0 & 58.1 & 57.7 & 62.1 & 62.5 & 49.3 & 33.3 & 59.6
& 58.55 & 1.7$\times$ & 47\% \\

\bottomrule
\end{tabular}

\label{tab:detailed_results_on_lvb_part2}
\end{table*}


\subsection{Compression Strategy Analysis}
\label{app:compression_strategy}

To determine whether the benefit of DAGC simply comes from reducing the number
of graph nodes, we compare it with two content-agnostic compression strategies
under the same 47\% retained-node budget. Uniform Merge combines neighboring
chunks using a fixed pattern, while Random Merge constructs merged units without
using video content. All variants use Qwen2.5-VL-7B and are evaluated on
LongVideoBench.

\begin{table*}[h]
\centering
\caption{
Comparison of different compression strategies on LongVideoBench using
Qwen2.5-VL-7B under the same retained-node budget.
}
\tablesize
\setlength{\tabcolsep}{5pt}
\renewcommand{\arraystretch}{1.02}

\begin{tabular}{lcc}
\toprule
\textbf{Method} &
\textbf{Accuracy} &
\textbf{Retained} \\
\midrule
Random Merge  & 61.53 & 47\% \\
Uniform Merge & 62.28 & 47\% \\
\rowcolor{oursbg}
DAGC          & \textbf{63.07} & 47\% \\
\bottomrule
\end{tabular}

\label{tab:compression_strategy_appendix}
\end{table*}

As shown in Table~\ref{tab:compression_strategy_appendix}, DAGC achieves 63.07 accuracy,
outperforming Uniform Merge by 0.79 percentage points and Random Merge by
1.54 percentage points under the same graph budget. Therefore, the performance
of DAGC cannot be explained solely by generic node reduction. Local visual
similarity provides a simple but effective criterion for identifying redundant
adjacent regions. We further examine whether more explicit event-boundary
modeling improves this representation in
Appendix~\ref{app:event_boundary_analysis}.


\subsection{Additional Event-Boundary Analysis}
\label{app:event_boundary_analysis}

The main paper evaluates learned EfficientGEBD boundaries as hard constraints
within DAGC and shows that they do not provide a consistent overall QA
improvement. Here, we provide an additional experiment with richer boundary
signals and further discuss the distinction between event segmentation and
redundancy-aware graph compression.

\paragraph{Hybrid boundary signals.}
In addition to learned event boundaries, we construct a lightweight
boundary-aware variant that combines appearance changes, motion changes, and
subtitle-semantic changes. We evaluate this variant on the complete MLVU
Needle subset while keeping the downstream retrieval and reasoning pipeline
unchanged.

\begin{table*}[h]
\centering
\caption{
Comparison of DAGC with a hybrid boundary-aware variant on the complete
MLVU Needle subset. The hybrid variant combines appearance, motion, and
subtitle-semantic boundary signals. The accuracy difference is not
statistically significant under a paired McNemar test ($p=0.25$).
Graph Construction Time reports the cumulative compute time summed across all GPUs, whereas Wall Time denotes the actual elapsed wall-clock time.
}
\tablesize
\setlength{\tabcolsep}{5pt}
\renewcommand{\arraystretch}{1.03}

\begin{tabular}{lccc}
\toprule
\textbf{Method} &
\textbf{Accuracy} &
\shortstack{\textbf{Graph Construction}\\\textbf{Time (s)}} &
\shortstack{\textbf{Wall}\\\textbf{Time (s)}} \\
\midrule

DAGC
& \textbf{82.25} (292/355)
& 30,741
& 13,780 \\

Hybrid Boundary DAGC
& 81.41 (289/355)
& 61,477
& 24,061 \\

\bottomrule
\end{tabular}

\label{tab:hybrid_boundary}
\end{table*}

As shown in Table~\ref{tab:hybrid_boundary}, incorporating richer boundary
signals changes accuracy from 82.25\% to 81.41\%. The difference is not
statistically significant ($p=0.25$), while graph-construction time nearly
doubles and the total wall-clock time increases substantially. Thus, richer
boundary cues do not provide a favorable accuracy--efficiency trade-off in
this setting.

\paragraph{Event segmentation versus DAGC.}
Event segmentation and DAGC optimize different objectives. Event segmentation seeks perceptually or semantically coherent temporal partitions, whereas DAGC aims to reduce redundant graph-construction units while preserving access to question-relevant evidence. Consequently, generic event boundaries need not align with QA evidence: a perceptual transition may be irrelevant to a question, while a brief object-state change, subtitle, or visual detail within a longer event may be decisive.

Event-based graph construction also introduces additional cost through full-video boundary inference and potentially more graph nodes and retrieval candidates. Moreover, variable-duration events still require fixed-budget or sparse frame sampling before LVLM processing.

In contrast, DAGC directly merges adjacent redundant chunks with a bounded span while retaining their original indices. Retrieved super-nodes can therefore be mapped back to fine-grained evidence for reranking and reasoning. These observations explain why more structured event partitions do not necessarily improve downstream QA: for long-video RAG, preserving retrievable fine-grained evidence is more important than enforcing semantically complete event boundaries.

\subsection{Component Ablation}
\label{app:component_ablation}

We further investigate the contribution of individual components of DAGC on
LongVideoBench. Tables~\ref{tab:component_ablation_lvb_part1}
and~\ref{tab:component_ablation_lvb_part2} compare the complete pipeline with
variants that remove super-node compression (SN), original-chunk recovery and
reranking (RR), or temporal expansion (TE).

Among the compressed variants, the complete DAGC pipeline obtains the highest
overall accuracy of 63.1. Removing original-chunk recovery and reranking causes
the largest degradation, reducing overall accuracy to 60.6, which highlights
the importance of recovering fine-grained evidence after coarse graph
retrieval. Temporal expansion also contributes to final performance, while
super-node compression is primarily responsible for the efficiency gain.

This result is consistent with the event-boundary analysis in
Appendix~\ref{app:event_boundary_analysis}: modifying the temporal partition
alone does not consistently improve final QA, whereas recovering and reranking
precise original evidence has a substantially larger effect.

\begin{table*}[h]
\centering
\caption{
Component ablation results on LongVideoBench, Part I.
The table reports the first group of question-category accuracies.
SN denotes super-node compression, RR denotes original-chunk recovery and
reranking, and TE denotes temporal expansion.
}
\tablesize
\setlength{\tabcolsep}{2.4pt}

\begin{tabular}{lccccccccc}
\toprule
\textbf{Variant} &
\textbf{E2O} &
\textbf{SSS} &
\textbf{T2E} &
\textbf{S2O} &
\textbf{SAA} &
\textbf{TAA} &
\textbf{S2A} &
\textbf{SOS} &
\textbf{T2A} \\
\midrule

Qwen2.5-VL-7B
& 70.8 & 46.4 & 66.2 & 55.6 & 62.8 & 53.7 & 72.7 & 63.0 & 59.3 \\

+ Vgent
& 75.4 & 46.4 & 72.3 & 66.7 & 61.1 & 59.8 & 72.1 & 65.4 & 70.4 \\

+ DAGC w/o SN
& 72.3 & 43.8 & 70.8 & 65.3 & 56.3 & 57.3 & 70.1 & 65.0 & 67.9 \\

+ DAGC w/o RR
& 72.3 & 42.7 & 72.3 & 63.9 & 58.3 & 57.3 & 70.1 & 64.2 & 69.1 \\

+ DAGC w/o TE
& 75.4 & 44.8 & 69.2 & 66.7 & 58.3 & 58.5 & 70.5 & 65.4 & 67.9 \\

\rowcolor{oursbg}
+ DAGC
& 75.4 & 44.8 & 69.2 & 68.1 & 58.3 & 58.5 & 70.5 & 65.4 & 67.9 \\

\bottomrule
\end{tabular}

\label{tab:component_ablation_lvb_part1}
\end{table*}

\begin{table*}[h]
\centering
\caption{
Component ablation results on LongVideoBench, Part II.
The table reports the remaining question-category accuracies together with
overall accuracy, relative wall-clock speedup, and retained-node ratio.
The full DAGC variant achieves the best overall accuracy among the compressed
variants while preserving the efficiency advantage over Vgent.
}
\tablesize
\setlength{\tabcolsep}{2.4pt}

\begin{tabular}{lcccccccc|ccc}
\toprule
\textbf{Variant} &
\textbf{S2E} &
\textbf{T3O} &
\textbf{T2O} &
\textbf{O3O} &
\textbf{O2E} &
\textbf{T3E} &
\textbf{TOS} &
\textbf{E3E} &
\shortstack{\textbf{Overall}\\\textbf{Accuracy}} &
\shortstack{\textbf{Wall}\\\textbf{Speedup}} &
\textbf{Retained} \\
\midrule

Qwen2.5-VL-7B
& 71.0 & 58.1 & 57.7 & 62.1 & 62.5 & 49.3 & 33.3 & 59.6
& 59.4 & 1.0$\times$ & -- \\

+ Vgent
& 76.3 & 59.5 & 64.1 & 60.6 & 69.3 & 49.3 & 40.0 & 68.1
& 63.3 & 1.0$\times$ & 100\% \\

+ DAGC w/o SN
& 75.3 & 58.1 & 67.9 & 66.7 & 69.3 & 50.7 & 40.0 & 68.1
& 62.6 & 1.0$\times$ & 100\% \\

+ DAGC w/o RR
& 74.2 & 58.1 & 65.4 & 62.1 & 65.5 & 49.3 & 38.7 & 67.0
& 60.6 & 1.6$\times$ & 47\% \\

+ DAGC w/o TE
& 73.1 & 55.4 & 67.9 & 65.2 & 67.0 & 48.6 & 41.3 & 65.9
& 62.2 & 1.6$\times$ & 47\% \\

\rowcolor{oursbg}
+ DAGC
& 74.2 & 56.8 & 69.2 & 68.2 & 68.2 & 49.3 & 42.7 & 67.0
& 63.1 & 1.6$\times$ & 47\% \\

\bottomrule
\end{tabular}

\label{tab:component_ablation_lvb_part2}
\end{table*}


\subsection{Parameter Sensitivity}
\label{app:parameter_sensitivity}

We study the sensitivity of DAGC to the visual-similarity threshold
$\tau_v$, the maximum merging window $W$, and the number of initial retrieval
seeds top-$k_s$. Tables~\ref{tab:parameter_sensitivity_lvb_part1}
and~\ref{tab:parameter_sensitivity_lvb_part2} report the complete
category-level results on LongVideoBench.

The default configuration
($\tau_v=0.95$, $W=3$, top-$k_s=12$)
achieves the highest overall accuracy of 63.1 among the tested settings while
retaining only 47\% of the original graph nodes. More aggressive compression
further reduces the number of nodes but gradually increases the risk of losing
fine-grained temporal evidence.

\begin{table*}[h]
\centering
\caption{
Parameter sensitivity results on LongVideoBench, Part I.
The table reports the first group of question-category accuracies under
different adaptive graph compression settings.
}
\tablesize
\setlength{\tabcolsep}{2.4pt}

\begin{tabular}{lccccccccc}
\toprule
\textbf{Setting} &
\textbf{E2O} &
\textbf{SSS} &
\textbf{T2E} &
\textbf{S2O} &
\textbf{SAA} &
\textbf{TAA} &
\textbf{S2A} &
\textbf{SOS} &
\textbf{T2A} \\
\midrule

$\tau_v$=0.95, $W$=3, top-$k_s$=12 (Default)
& 75.4 & 44.8 & 69.2 & 68.1 & 58.3 & 58.5 & 70.5 & 65.4 & 67.9 \\

$\tau_v$=0.95, $W$=3, top-$k_s$=5
& 71.9 & 45.4 & 70.8 & 65.3 & 56.9 & 58.5 & 69.0 & 64.2 & 69.1 \\

$\tau_v$=0.95, $W$=3, top-$k_s$=20
& 73.8 & 43.6 & 72.3 & 65.3 & 56.9 & 57.3 & 69.3 & 63.0 & 67.9 \\

$\tau_v$=0.95, $W$=1, top-$k_s$=12
& 72.3 & 43.8 & 70.8 & 65.3 & 56.3 & 57.3 & 70.1 & 65.0 & 67.9 \\

$\tau_v$=0.90, $W$=3, top-$k_s$=12
& 73.8 & 48.5 & 69.2 & 66.7 & 58.3 & 57.3 & 70.5 & 63.0 & 67.9 \\

$\tau_v$=0.95, $W$=7, top-$k_s$=12
& 72.3 & 45.4 & 70.8 & 65.3 & 58.3 & 57.3 & 70.5 & 64.2 & 70.4 \\

$\tau_v$=0.80, $W$=3, top-$k_s$=12
& 72.3 & 43.3 & 72.3 & 65.3 & 58.3 & 57.3 & 69.3 & 64.2 & 67.9 \\

\bottomrule
\end{tabular}

\label{tab:parameter_sensitivity_lvb_part1}
\end{table*}

\clearpage
\begin{table*}[h]
\centering
\caption{
Parameter sensitivity results on LongVideoBench, Part II.
The table reports the remaining question-category accuracies together with
overall accuracy, relative wall-clock speedup, and retained-node ratio.
The default setting achieves the best overall accuracy while maintaining
a clear efficiency gain.
}
\tablesize
\setlength{\tabcolsep}{2.4pt}

\begin{tabular}{lcccccccc|ccc}
\toprule
\textbf{Setting} &
\textbf{S2E} &
\textbf{T3O} &
\textbf{T2O} &
\textbf{O3O} &
\textbf{O2E} &
\textbf{T3E} &
\textbf{TOS} &
\textbf{E3E} &
\shortstack{\textbf{Overall}\\\textbf{Accuracy}} &
\shortstack{\textbf{Wall}\\\textbf{Speedup}} &
\textbf{Retained} \\
\midrule

$\tau_v$=0.95, $W$=3, top-$k_s$=12 (Default)
& 74.2 & 56.8 & 69.2 & 68.2 & 68.2 & 49.3 & 42.7 & 67.0
& 63.1 & 1.6$\times$ & 47\% \\

$\tau_v$=0.95, $W$=3, top-$k_s$=5
& 74.2 & 58.1 & 65.4 & 66.7 & 68.2 & 49.3 & 45.3 & 68.1
& 62.7 & 1.6$\times$ & 47\% \\

$\tau_v$=0.95, $W$=3, top-$k_s$=20
& 74.2 & 56.8 & 67.9 & 68.2 & 69.0 & 49.3 & 42.7 & 69.1
& 62.7 & 1.6$\times$ & 47\% \\

$\tau_v$=0.95, $W$=1, top-$k_s$=12
& 75.3 & 58.1 & 67.9 & 66.7 & 69.3 & 50.7 & 40.0 & 68.1
& 62.6 & 1.0$\times$ & 100\% \\

$\tau_v$=0.90, $W$=3, top-$k_s$=12
& 74.2 & 58.1 & 64.9 & 66.7 & 69.3 & 47.9 & 41.3 & 69.1
& 62.8 & 1.7$\times$ & 39\% \\

$\tau_v$=0.95, $W$=7, top-$k_s$=12
& 74.2 & 58.1 & 65.4 & 66.7 & 68.2 & 49.3 & 42.7 & 67.0
& 62.5 & 1.7$\times$ & 34\% \\

$\tau_v$=0.80, $W$=3, top-$k_s$=12
& 73.1 & 58.1 & 66.7 & 65.2 & 69.3 & 49.3 & 44.0 & 68.1
& 62.5 & 1.7$\times$ & 35\% \\

\bottomrule
\end{tabular}

\label{tab:parameter_sensitivity_lvb_part2}
\end{table*}

\subsection{Effects of Similarity Threshold and Compression Span}
\label{app:compression_parameter_effects}

\begin{figure*}[h]
    \centering

    \begin{minipage}[t]{0.47\linewidth}
        \centering
        \includegraphics[
            width=\linewidth,
            trim=0 0 0 0,
            clip
        ]{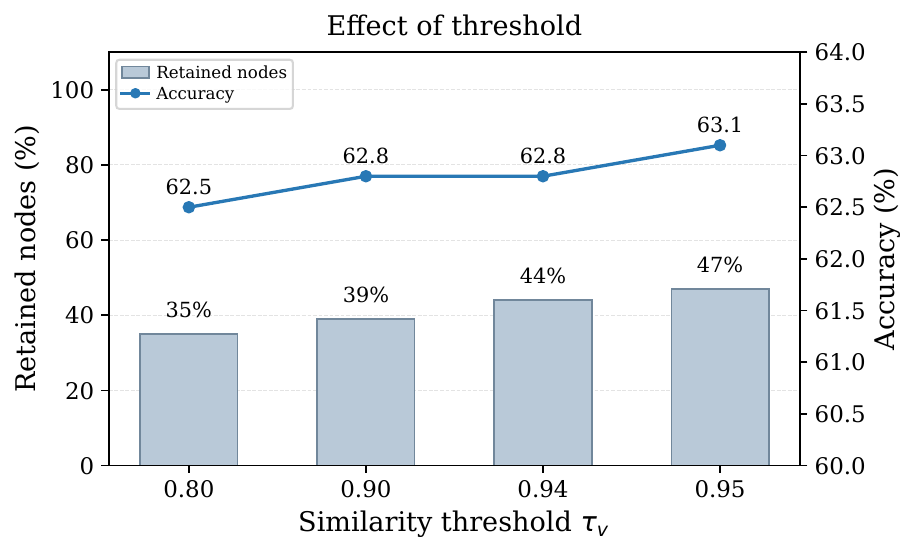}

        \vspace{-0.3em}
        {\small (a) Effect of similarity threshold $\tau_v$.}
    \end{minipage}
    \hfill
    \begin{minipage}[t]{0.47\linewidth}
        \centering
        \includegraphics[
            width=\linewidth,
            trim=0 0 0 0,
            clip
        ]{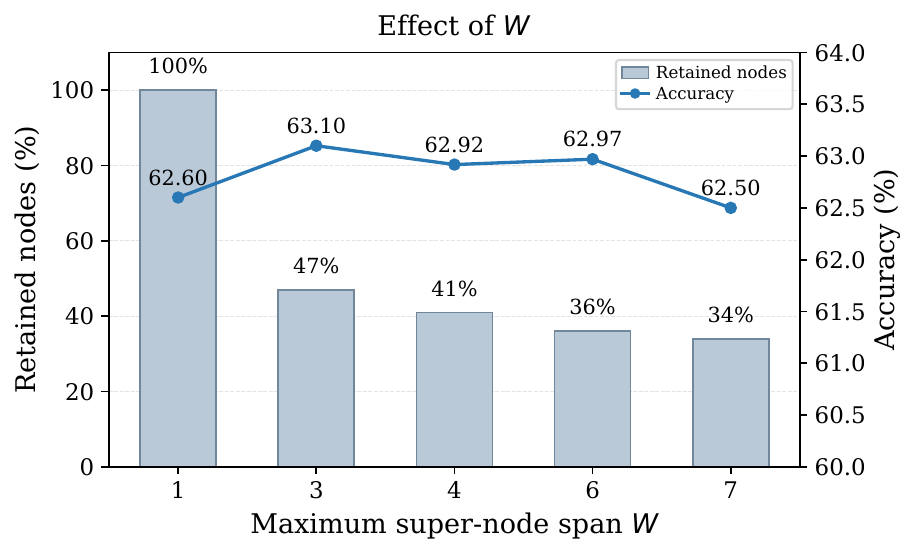}

        \vspace{-0.3em}
        {\small (b) Effect of maximum super-node span $W$.}
    \end{minipage}

    \caption{
    Effect of index-coarsening parameters on LongVideoBench. Bars show retained graph nodes, and lines show QA accuracy. (a) Higher $\tau_v$ yields more conservative merging and retains more nodes. (b) Larger super-node spans enable stronger coarsening but may reduce accuracy.
    }
    \label{fig:compression_parameter_effects}
\end{figure*}

To more directly visualize the accuracy--compression trade-off, we
independently vary the similarity threshold and the maximum super-node span
on LongVideoBench, as shown in Figure~\ref{fig:compression_parameter_effects}. Increasing $\tau_v$ from 0.80 to 0.95 makes the merging criterion more
conservative, increasing the retained-node ratio from 35\% to 47\%, while
accuracy improves from 62.5 to 63.1. This indicates that retaining additional
boundaries provides a small but consistent benefit when the similarity threshold is increased. The maximum super-node span exhibits a similar trade-off. A moderate span of $W=3$ achieves the highest accuracy of 63.1 while retaining 47\% of the graph nodes. Increasing $W$ further reduces the graph to 41\%, 36\%, and 34\% of its original size for $W=4$, 6, and 7, respectively. Performance remains relatively stable for moderate compression but decreases to 62.5 at $W=7$, indicating that overly long merging windows may remove useful fine-grained temporal structure.

\section{Limitation}

DAGC uses adjacent visual similarity as a lightweight proxy for local redundancy, which may fail to capture semantic changes in visually stable regions, such as evolving dialogue or subtle object-state transitions. Its effectiveness also depends on the degree of index coarsening: information-dense videos may require more conservative thresholds or merging spans. In addition, DAGC is designed for efficient long-video QA rather than precise event segmentation, and its super-nodes are not guaranteed to correspond to complete semantic events. Although we validate DAGC across multiple LVLM backbones and video RAG pipelines, broader evaluation on additional graph structures and retrieval frameworks remains future work.